%% file: CameraReady2027.tex
\documentclass[letterpaper]{article}
\usepackage[preprint]{aaai2027}
\usepackage[hyphens]{url}
\usepackage{graphicx}
\usepackage{natbib}
\usepackage{caption}
\usepackage{booktabs}
\usepackage{makecell}
\usepackage{amsmath}
\usepackage{amssymb}

\newcommand{\method}{SAM-Radar}
\newcommand{\benchmark}{RGBTR-Motion}

\title{Segment Any Motion with Radar: Robust Multimodal Moving-Object Segmentation and Tracking}
\author{
    Jue Wang\textsuperscript{1,2},
    Xuan Wang\textsuperscript{1}\corresponding,
    Hao Zhou\textsuperscript{2},
    Ruixiang Zhou\textsuperscript{3},\\
    Yixuan Zhou\textsuperscript{1,2},
    Tianshuo Yuan\textsuperscript{3},
    Jieming Ma\textsuperscript{1,2},\\
    Jie Zhang\textsuperscript{2},
    Fei Luo\textsuperscript{2}\corresponding
}
\affiliations{
    \textsuperscript{1}Harbin Institute of Technology, Shenzhen, China\\
    \textsuperscript{2}Great Bay University, Dongguan, China\\
    \textsuperscript{3}Shenzhen Institutes of Advanced Technology, Chinese Academy of Sciences, Shenzhen, China
}

\begin{document}
\maketitle

\input{sections/00_abstract}
\input{sections/01_introduction}
\input{sections/02_related_work}
\input{sections/03_benchmark}
\input{sections/04_method}
\input{sections/05_experiments}
\input{sections/06_conclusion}

\bibliography{aaai2027}
\end{document}

%% file: sections/00_abstract.tex
\begin{abstract}
Moving-object perception must decide which image regions correspond to real motion and keep every instance identified over time. Methods that read motion from appearance, optical flow, or estimated trajectories lose that evidence under poor illumination, adverse weather, reflections, and occlusion. Radar is a natural remedy because it measures radial velocity directly instead of inferring it from the photometric correspondence. However, existing benchmarks do not jointly provide radar measurements, dense moving-instance masks, and temporally consistent identities for surveillance. We therefore introduce \benchmark, a synchronized and calibrated fixed-camera benchmark that pairs RGB, thermal, and radar streams with dense instance masks and temporally consistent identities across diverse surveillance scenes. We also develop \method, a RGB, thermal and radar-based segmentation and tracking framework built on SAM~3.
\method{}'s radar-aware detector fuses calibrated RGBT features with radar returns that are grounded at their projected image locations, and motion supervision, implemented as foreground classification of those projected returns, teaches the detector to reject clutter without any text prompt. The tracker associates accepted radar returns with individual trajectories and uses them as physical evidence that a visually degraded target remains present. This allows it to bridge short periods of low visibility or occlusion and reconnect a reappearing target to its existing identity instead of starting a new track.
\method{} attains 0.7027 IoU and 0.8090 F1$_{50}$, and raises MOTA, HOTA, and IDF1 by 0.2977, 0.1603, and 0.2857 over the strongest competing values.


\end{abstract}

%% file: sections/01_introduction.tex
\section{Introduction}
\label{sec:introduction}

\begin{figure*}[t]
    \centering
    \includegraphics[width=0.95\textwidth]{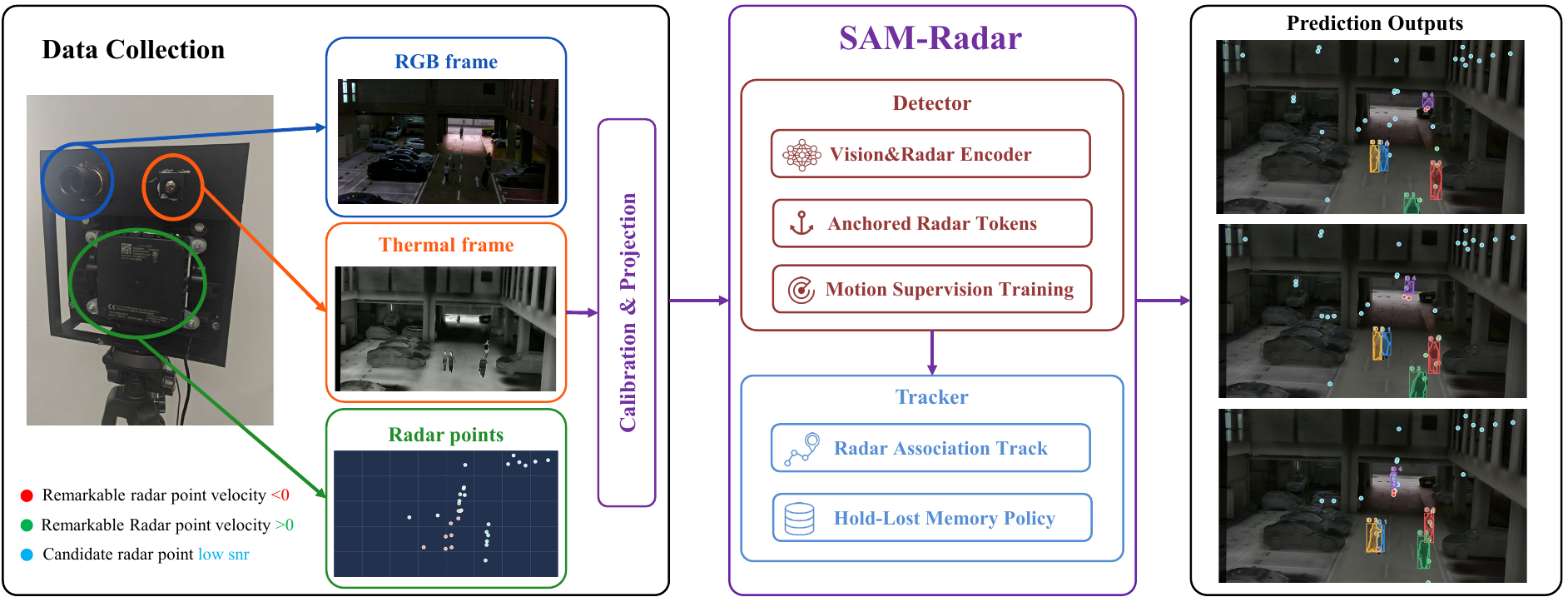}
    \caption{Overview of the acquisition and \method{} pipeline. A rigid rig records calibrated RGB, thermal, and radar streams. The detector fuses visual features with spatially anchored radar tokens trained by motion supervision, and the tracker combines radar association with Hold-Lost memory to produce temporally consistent masks and identities.}
    \label{fig:teaser}
\end{figure*}

Reliable moving-object perception must recover accurate masks and preserve the identities of physical movers over time. In fixed-camera surveillance, weak appearance corrupts boundaries, while crossings and occlusions disrupt identity over long recordings. We target physical agents, predominantly pedestrians, cyclists, and vehicles, rather than photometric or background changes. The task is therefore stricter than generic change detection but broader than closed-vocabulary detection.


Most established approaches derive motion from images. Optical-flow methods group short-term displacements or learn masks from flow fields~\cite{xie2022oclr,teed2020raft}; OCLR~\cite{xie2022oclr}, for example, maintains layered object masks through mutual occlusion. SegAnyMo~\cite{huang2025seganymotion} instead classifies long-range visual trajectories and prompts SAM~2~\cite{ravi2024sam2} to recover dense masks. Yet their motion evidence still comes from visual correspondence: illumination changes corrupt photometric evidence, thermal crossover reduces contrast, occlusion removes correspondences, and fast or subtle motion corrupts estimated tracks.

Camera and radar fusion is well established in autonomous driving~\cite{caesar2020nuscenes,nabati2021centerfusion,kim2023craft}, waterways~\cite{guan2024asyvrnet}, and other platform-centric perception. Our contribution is therefore not the first RGBT-radar dataset, but a surveillance-specific viewpoint, motion-centric task, and dense temporal annotation. Mobile-platform datasets emphasize ego-centric geometry and category-level perception; fixed-camera surveillance instead requires long-duration observation from a stationary viewpoint, together with pixel-level masks and persistent identities. Existing multimodal corpora do not directly support this setting.

Radar measures motion rather than inferring it from photometric correspondence, so radial velocity can remain informative when appearance is weak. Radar alone is nevertheless sparse and noisy, and its velocity cue vanishes when a target stops or moves tangentially to the line of sight. It therefore complements rather than replaces appearance. When visual evidence becomes unreliable, associated radar measurements remain available. Thermal imagery complements RGB by providing clearer boundaries in low light. The central challenge is to ground sparse radar evidence in the image and assign it consistently to object tracks.

We build \method{} (\textbf{Segment Any Motion with Radar}) on SAM~3's decoupled detector and tracker~\cite{carion2025sam3} and train them in two separate stages. In Stage 1, the detector combines complementary RGB and thermal features with radar returns anchored to their projected image neighborhoods, allowing it to locate movers using both visual structure and measured motion.
Motion supervision teaches the detector to distinguish target returns from multipath reflections and background clutter, reducing clutter-induced predictions. In Stage 2, the detector and radar encoder are frozen, and the tracker matches each return to at most one active track based on spatial, motion, and identity consistency. Confident matches reinforce the corresponding track when visual evidence weakens. To bridge short occlusions, Hold-Lost memory preserves the last reliable visual state while radar continues to support the target and releases the identity only after both visual and radar evidence remain absent.


We also organize \benchmark, a synchronized and calibrated fixed-camera benchmark with 107 sequences and 8,537 annotated frames across daytime, nighttime, rainy, and indoor conditions. It provides frame-level masks and persistent identities, with an acquisition-grouped train/test split that prevents temporal leakage. Representative motion-segmentation and multimodal baselines are adapted to the same data and evaluation protocol.

Our contributions are threefold:
\begin{itemize}
    \item We construct \benchmark, a synchronized and calibrated fixed-camera surveillance benchmark with dense masks and persistent identities, addressing a task and annotation setting underrepresented by existing automotive-oriented tri-modal data.
    \item We introduce \method, whose radar-aware detector combines calibrated RGBT features with spatially anchored radar tokens, while tracker uses radar association, confidence-gated radar residual injection, and Hold-Lost memory to preserve identities through temporary appearance failures.
    \item We develop a radar-aware two-stage training strategy. Stage 1 trains the detector to discover and segment moving objects from the aligned RGBT-radar inputs. Stage 2 freezes the detector and trains the tracker to preserve object identities over time, especially when visual evidence weakens.
\end{itemize}


%% file: sections/02_related_work.tex
\section{Related Work}
\label{sec:related}

\subsection{Moving-Object Segmentation}
Classical motion segmentation groups pixels or trajectories by geometric consistency~\cite{brox2010object,ochs2014segmentation}. In fixed-camera surveillance, background subtraction remains vulnerable to nuisance appearance changes~\cite{stauffer1999adaptive,zivkovic2004improved,barnich2011vibe,bouwmans2014traditional}, as systematically benchmarked by CDnet~\cite{wang2014cdnet}. Modern systems instead use optical flow~\cite{teed2020raft}; OCLR~\cite{xie2022oclr}, for example, predicts depth-ordered object layers and learns temporal shape consistency from synthetic compositions.

Instance-level video tasks such as MOTS~\cite{voigtlaender2019mots} and video instance segmentation~\cite{yang2019vis} couple masks with identities but assume a closed vocabulary. SAM~\cite{kirillov2023sam} established promptable segmentation, SAM~2~\cite{ravi2024sam2} introduced streaming video memory, and SAM~3~\cite{carion2025sam3} unified open-vocabulary detection and tracking. SegAnyMo~\cite{huang2025seganymotion} classifies long-range point tracks and prompts SAM~2 for dense masks. Motion-oriented uses of these models still rely on visual correspondence and fail when that evidence disappears. \method{} instead uses measured radar motion, spatially anchored radar tokens, and Hold-Lost memory.

\subsection{Vision-Radar Perception}
RGB supplies texture and color, while thermal imagery complements it in low light~\cite{hwang2015kaist,li2019rgbt234,li2021lasher}. Condition-aware fusion improves robustness when either stream degrades~\cite{broedermann2025cafuser}, but both remain appearance based. Radar adds independent range and radial velocity, motivating camera-radar systems for driving and corresponding adverse-weather and 4D-radar datasets~\cite{caesar2020nuscenes,nabati2021centerfusion,kim2023craft,sheeny2021radiate,palffy2022vod,paek2022kradar}.

ASY-VRNet~\cite{guan2024asyvrnet} combines visual features and 4D radar maps for mobile waterway perception. RADCI~\cite{radci} is a fixed-tripod RGBT-radar benchmark for detection and tracking, with 2D boxes and target IDs. These works establish the value of multimodal fusion, but neither provides motion-centric dense instance masks with sequence-level identity annotations. We instead align thermal features and sparse returns to the visible image, using radar as track-specific motion evidence.


%% file: sections/03_benchmark.tex
\section{The \benchmark{} Benchmark}
\label{sec:benchmark}

\begin{figure}[t]
    \centering
    \includegraphics[width=0.46\textwidth]{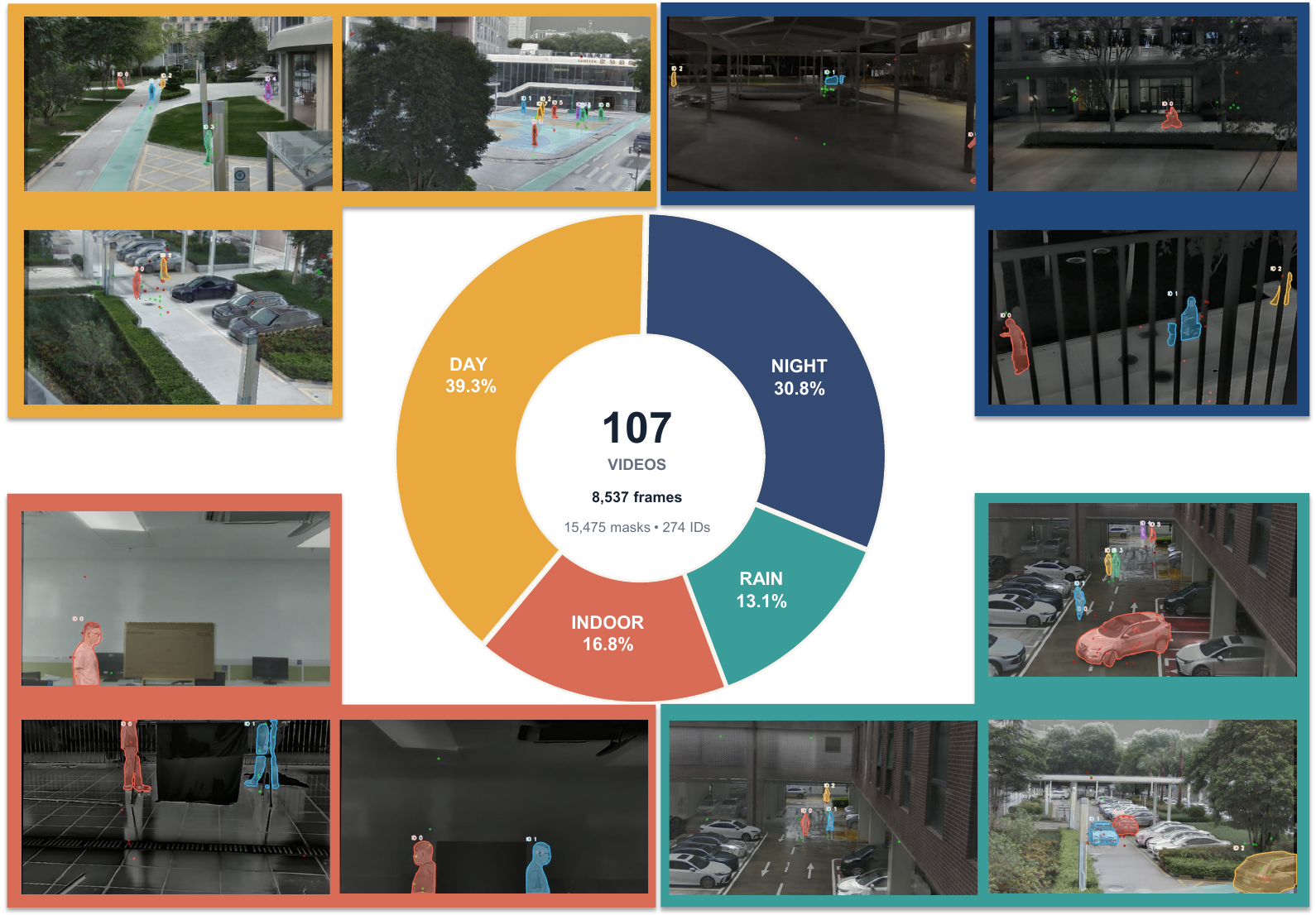}
    \caption{Composition of the benchmark \benchmark{}. Example panels show calibrated RGBT fusion, sequence-local moving-instance masks and identities, and projected radar returns for daytime, nighttime, rainy, and indoor monitoring conditions.}
    \label{fig:datasets}
\end{figure}
\subsection{Data Collection}

\benchmark{} is designed specifically for motion perception from stationary surveillance viewpoints. Data are acquired with a rigid tri-modal rig comprising a $1280\!\times\!720$ visible-light camera, a $640\!\times\!512$ long-wave infrared camera, and a ZF 4D millimeter-wave radar with 12 transmit and 16 receive channels. Both cameras and radar record at 10~fps. All three sensors are mounted on a common plate and remain fixed throughout each acquisition, providing consistent geometry across daytime, nighttime, rainy, and indoor recordings. Complementing the aligned visual streams, the radar records each return’s 3D position, range-related geometry, radial velocity, signal-to-noise ratio (SNR), and timestamp. 

The benchmark targets the sustained, scene-traversing movers described in Section~\ref{sec:introduction}, chiefly pedestrians, cyclists, and vehicles, and preserves identity through surveillance events such as entry, target crossing, occlusion, and exit. The audited benchmark contains 107 sequences, 8,537 annotated frames, 15,475 instance masks, and 274 sequence-local identities.
Table~\ref{tab:benchmark_comparison} focuses on RADCI because it provides the closest sensor and acquisition setting. RADCI reports seven road and square scenarios recorded by day and night. \benchmark{} expands the coverage to outdoor and indoor surveillance spanning daytime, nighttime, and rain. More importantly, it goes beyond category-level box annotation by linking motion-centric instance masks to sequence-local identities, enabling pixel-level segmentation and tracking through entry, crossing, occlusion, and exit.
\begin{table}[t]
    \centering
    \caption{Comparison with RADCI~\cite{radci}, the closest synchronized RGBT-radar benchmark. The two share a sensor configuration; \benchmark{} differs in condition coverage, annotation granularity, and how a target is defined.}
    \label{tab:benchmark_comparison}
    \small
    \setlength{\tabcolsep}{4pt}
    \renewcommand{\arraystretch}{1.15}
    \begin{tabular}{@{}p{0.18\columnwidth}p{0.38\columnwidth}p{0.38\columnwidth}@{}}
        \toprule
                     & RADCI    & \benchmark{} (ours) \\
        \midrule
        Modalities   & RGB, thermal, radar    & RGB, thermal, radar \\
        Environment  & road, square & road, gym, square, indoor\\
        Conditions   & Day, night             & Day, night, rain \\
        Sequences    & 94                     & 107 \\
        Annotation   & 2D boxes with IDs      & Instance masks with IDs \\
        Target & Predefined categories  & Any sustained mover \\
        Task         & Detection, tracking    & Segmentation, tracking \\
        \bottomrule
    \end{tabular}
\end{table}
Figure~\ref{fig:datasets} visualizes this composition together with calibrated RGBT-radar surveillance examples.

\begin{figure*}[t]
    \centering
    \includegraphics[width=0.95\textwidth]{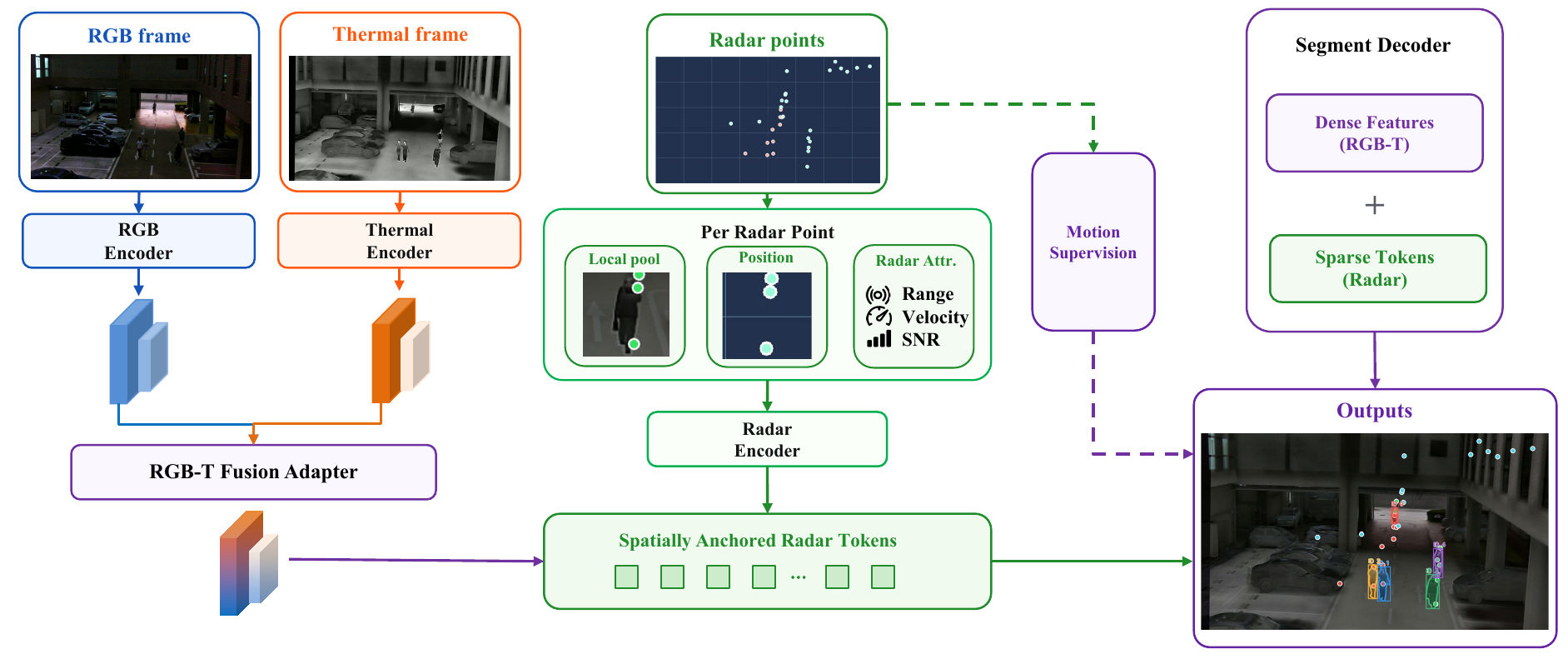}
    \caption{For each projected radar return, the radar-aware detector combines local RGBT context, image position, and radar attributes to form a spatially anchored token. Motion supervision teaches it to distinguish target-associated returns from clutter, and the segmentation decoder combines these sparse tokens with dense visual features to predict moving instances.}
    \label{fig:method}
\end{figure*}

\subsection{Temporal Synchronization and Spatial Calibration}
All sensor streams are mapped to the visible-camera timeline. Visible and infrared frames are undistorted, and infrared frames are warped into visible-image coordinates using the calibrated cross-camera transformation. For a radar point $\mathbf{x}^{r} \in \mathbb{R}^{3}$, calibrated rotation $\mathbf{R}_{rv}$ and translation $\mathbf{t}_{rv}$ transform it into the visible-camera coordinate system:
\begin{equation}
    \mathbf{x}^{v}=\mathbf{R}_{rv}\mathbf{x}^{r}+\mathbf{t}_{rv}.
\end{equation}
The visible camera matrix and distortion model then project $\mathbf{x}^{v}$ to pixel coordinates $(u,v)$. Returns outside the image plane are discarded. Video frame $t$ is paired with the $t$-th radar timestamp and groups returns in a small temporal neighborhood. We retain reliable moving points together with high-SNR static or slow points, then convert projected coordinates into the undistorted visible-image system, producing
\begin{equation}
    \mathcal{R}_{t}=\{(u_j,v_j,d_j,\dot d_j,s_j)\}_{j=1}^{N_t},
\end{equation}
where $d_j$, $\dot d_j$, and $s_j$ denote range, radial velocity, and SNR, respectively.

\subsection{Annotations and Tasks}
 Each visible frame is annotated in COCO-compatible format with instance masks, boxes, categories, and sequence-local persistent identities. The labels are motion-centric rather than restricted to a predefined surveillance object taxonomy. They support two tasks under a common annotation set: frame-level moving-object instance segmentation and video-level segmentation and tracking. 
 The annotation policy covers temporary full occlusion, targets leaving the field of view, and stationary intervals of otherwise moving objects.



%% file: sections/04_method.tex
\section{Method}
\label{sec:method}

\subsection{Overview}
\begin{figure*}[t]
    \centering
    \includegraphics[width=0.95\textwidth]{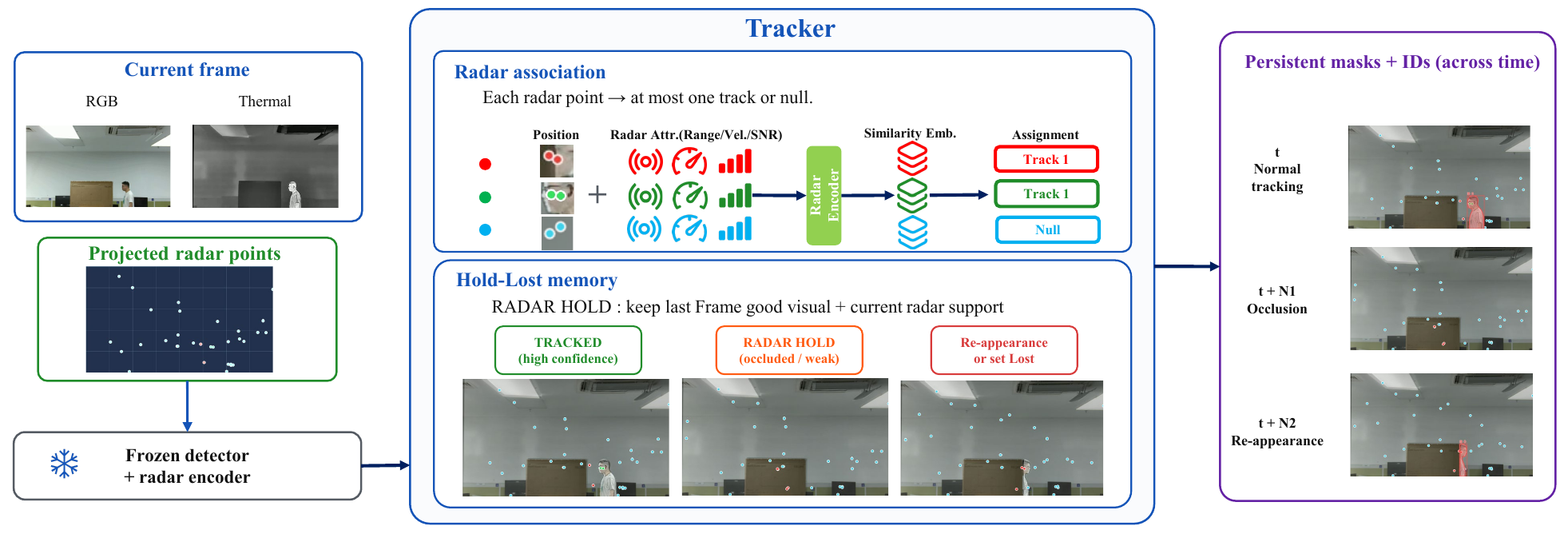}
    \caption{The radar-conditioned tracker receives the current RGBT observations and projected radar returns. Radar association combines projected position, radar attributes, and descriptor similarity to assign each return to at most one active track. When visual evidence weakens but radar continues to support the target, Hold-Lost memory preserves the last reliable visual state.}
    \label{fig:tracker}
\end{figure*}

Given synchronized RGB frames $I^R_{1:T}$, aligned thermal frames $I^T_{1:T}$, and projected radar sets $\mathcal{R}_{1:T}$, the model predicts instance masks and persistent identities for physical movers. We retain the decoupled detector and memory tracker of SAM~3~\cite{carion2025sam3} and train them in two stages. The detector first learns frame-level discovery and segmentation from all three modalities. Its weights and radar encoder are then frozen while the tracker learns radar association, feature updates, and Hold-Lost memory. This separation prevents temporal training from changing the learned object discovery process. Figures~\ref{fig:method} and~\ref{fig:tracker} show the data flow.

\subsection{Radar-Aware Multimodal Detector}
\label{sec:detector}

\paragraph{Calibrated RGBT fusion}
Directly stacking thermal and RGB images at the input assumes compatible channel statistics and exposes the pretrained RGB representation to thermal noise. We instead process the aligned images with separate vision trunks initialized from the same pretrained encoder. Let $F^R$ and $F^T$ denote their features, and let $A_R$ and $A_T$ be lightweight adapters. Their detector-level fusion is
\begin{equation}
\begin{aligned}
    \bar F^m &= A_m(F^m), \quad m\in\{R,T\},\\
    F^{RT} &= F^R+\Phi(\bar F^R,\bar F^T).
\end{aligned}
\end{equation}
Here, $\Phi$ is the learned fusion module. It predicts independent sigmoid channel weights from globally pooled adapted features, combines the weighted and spatial-fusion branches with a learned scalar initialized to $0.5$, and returns a residual. Consequently, $F^{RT}$ retains a direct RGB path while incorporating thermal evidence.

\paragraph{Spatially anchored radar tokens}
A direct approach is to encode only the projected coordinates and physical attributes of each radar return with an MLP. This retains range and radial velocity but omits the visual structure at the projected location. For return $r_j=(u_j,v_j,d_j,\dot d_j,s_j)$, let $p_j$ encode position $(u_j,v_j)$, let $\ell_j$ be the locally pooled RGBT feature, and let $f_j^{\mathrm{attr}}$ encode the normalized attributes $(d_j,\dot d_j,s_j)$. We construct its token as
\begin{equation}
\begin{aligned}
    b_j &= p_j+W_{\ell}\ell_j,\\
    g_j &= \sigma(W_g f_j^{\mathrm{attr}}),\\
    z_j &= \operatorname{LN}(b_j+g_j\odot f_j^{\mathrm{attr}}
    +e_{\mathrm{rad}}+e_j^{\mathrm{fg}}).
\end{aligned}
\end{equation}
$W_{\ell}$ and $W_g$ are learned projections, $e_{\mathrm{rad}}$ identifies the radar modality, and $e_j^{\mathrm{fg}}$ is a soft foreground embedding predicted for the return. The gate $g_j$ controls the attribute residual, while layer normalization produces the detector token $z_j$. This construction links each sparse physical measurement to its local image context.

\paragraph{Motion supervision}
Standard mask and box losses supervise the final object predictions but do not indicate which radar returns correspond to targets. We therefore assign label $y_j\in\{0,1\}$ to each projected return and predict its foreground probability $\hat y_j$. The corresponding focal objective~\cite{lin2017focal} and total detector objective are
\begin{equation}
\begin{aligned}
    q_j &= y_j\hat y_j+(1-y_j)(1-\hat y_j),\\
    \mathcal{L}_{\mathrm{rad}}
    &= -\frac{1}{N_t}\sum_{j=1}^{N_t}
    \beta_{y_j}(1-q_j)^{\gamma}\log q_j, \\
    \mathcal{L}_{\mathrm{det}}
    &= \mathcal{L}_{\mathrm{SAM3}}+\lambda_{\mathrm{rad}}\mathcal{L}_{\mathrm{rad}}.
\end{aligned}
\end{equation}
Here, $N_t$ is the number of returns, $\beta_{y_j}$ is the class weight, and $\gamma$ is the focal exponent. $\mathcal{L}_{\mathrm{SAM3}}$ collects the classification, box, generalized-IoU~\cite{rezatofighi2019giou}, and mask losses, while $\lambda_{\mathrm{rad}}$ weights motion supervision.

\subsection{Radar-Conditioned Tracker}
\label{sec:tracker}

SAM~3 uses separate feature pyramids for detection and tracking, so detector-side fusion does not expose the high-resolution tracking path to thermal input. We therefore fuse thermal features at every tracker scale while freezing the spatial encoder, detector, and memory writer; stage two updates only temporal multimodal components.


\paragraph{Radar association}
A hard previous-mask gate cannot recover after drift, whereas independent matching can assign one return to several tracks. We instead multiply four pairwise terms: spatial support $S_{ij}^t$, SNR reliability $Q_j^t$, range-velocity continuity $D_{ij}^t$, and descriptor similarity $C_{ij}^t$. They test whether a return is spatially close, reliable, consistent with the track's range-velocity history, and similar to its accumulated radar descriptor:
\begin{equation}
\begin{aligned}
    w_{ij}^t &= S_{ij}^t\cdot Q_j^t\cdot D_{ij}^t\cdot C_{ij}^t,\\
    a_{ij}^t &= \frac{w_{ij}^t}
    {w_{\varnothing j}^t+\sum_{k\in\mathcal{T}_t}w_{kj}^t},\\
    i_j^* &= \arg\max_{i\in\mathcal{T}_t}a_{ij}^t,\\
    j&\mapsto i_j^*\quad\text{if}\quad
    a_{i_j^*j}^t\geq\tau_{\mathrm{own}}.
\end{aligned}
\end{equation}
Here, $\mathcal{T}_t$ is the active-track set, $w_{\varnothing j}^t=0.25$ is the fixed null evidence for clutter, and $\tau_{\mathrm{own}}=0.35$ is the ownership threshold. Each return is assigned to at most one track and remains unassigned when the null option or confidence test rejects it.

Once a track has accepted a radar measurement, we call it \emph{radar initialized}. Its spatial prior is then floored at $0.05$ outside the previous visual box so that a displaced return can still compete for association. Accepted returns are pooled into a current measurement and a track-specific descriptor.

\paragraph{Radar feature updates}
Only observed returns update the persistent radar state. If a measurement is missing, the last descriptor and radial velocity are retained for at most four frames with decaying confidence, and range is extrapolated under constant radial velocity. Rendering this predicted state as a spatial return would reuse an old image location even though range extrapolation does not determine image-plane motion. We therefore allow only current measurements to populate the sparse spatial residual; propagated descriptors can affect the global object representation and pointer, but not the spatial radar map.

Let $M_i^t$ be the current sparse radar map, $h_i^t$ the current or propagated track descriptor, $c_i^t$ its confidence, $F_i^t$ the visual tracking feature, and $o_i^t$ the object pointer. Radar is injected after memory attention and before mask decoding as
\begin{equation}
\begin{aligned}
    \Delta F_i^t
    &=\tanh(g_s)A_s(M_i^t)+\tanh(g_p)A_p(h_i^t),\\
    \widetilde F_i^t &=F_i^t+c_i^t\Delta F_i^t,\\
    \widetilde o_i^t
    &=o_i^t+c_i^t\tanh(g_p)A_p(h_i^t).
\end{aligned}
\end{equation}
$A_s$ and $A_p$ are learned residual projections, and $g_s$ and $g_p$ are scalar gates. 

\paragraph{Hold-Lost memory}
Immediately discarding a visually weak track loses objects during short occlusions, whereas writing every low-confidence prediction into memory can corrupt later propagation. Hold-Lost memory uses three states to separate these cases. TRACKED denotes a visually reliable prediction and refreshes the most recent reliable frame. If visual confidence becomes unreliable while a current radar measurement remains associated, the track enters HOLD and keeps that reliable frame in the seven-slot memory selection window. Visual recovery returns the track to TRACKED. If neither vision nor current radar supports the track for four consecutive frames, it becomes LOST; its mask and score are suppressed and its active anchor is released.


%% file: sections/05_experiments.tex
\section{Experiments}

\label{sec:experiments}
\begin{table*}[t]
    \centering
    \caption{Main comparison on the \benchmark{}. IoU and F1$_{50}$ evaluate foreground and instance moving-object segmentation. MOTA, MOTP, HOTA, and IDF1 evaluate tracking. All models are fine-tuned on \benchmark{}; $^*$ denotes methods additionally adapted to accept aligned RGB, thermal, and radar
  inputs. Best results are in bold and second-best results are underlined.}
    \label{tab:main_results}
    \begin{tabular}{@{}llcccccc@{}}
        \toprule
        Method & Modalities & IoU $\uparrow$ & F1$_{50}$ $\uparrow$ & MOTA $\uparrow$ & MOTP $\uparrow$ & HOTA $\uparrow$ & IDF1 $\uparrow$ \\
        \midrule
        OCLR Hybrid~\cite{xie2022oclr} & RGB & 0.4669 & 0.5885 & 0.2603 & 0.7349 & 0.3583 & 0.3449 \\
        SegAnyMo DINO+SAM~\cite{huang2025seganymotion} & RGB & 0.6103 & 0.6070 & 0.3074 & 0.7568 & 0.4166 & \underline{0.4756} \\
        ASY-VRNet$^{*}$~\cite{guan2024asyvrnet} & RGB+T+R & 0.4884 & 0.4805 & 0.2990 & \textbf{0.7716} & 0.3390 & 0.2808 \\
        CAFuser$^{*}$~\cite{broedermann2025cafuser} & RGB+T+R & 0.6489 & 0.6111 & 0.3188 & 0.7650 & 0.4308 & 0.3750 \\
        RADCI-RCINet$^{*}$~\cite{radci} & RGB+T+R & \underline{0.6574} & \underline{0.6243} & \underline{0.3259} & \underline{0.7666} & \underline{0.4415} & 0.3931 \\
        
        \midrule
        \method{} (ours) & RGB+T+R & \textbf{0.7027} & \textbf{0.8090} & \textbf{0.6236} & 0.7559 & \textbf{0.6018} & \textbf{0.7613} \\
        \bottomrule
    \end{tabular}
\end{table*}

\subsection{Experimental Setup}
\paragraph{Baselines}
We evaluated five representative baselines using the same acquisition-grouped split and annotation protocol. OCLR Hybrid~\cite{xie2022oclr} and SegAnyMo DINO+SAM~\cite{huang2025seganymotion} retained their native RGB motion inputs and predicted class-agnostic moving-instance masks. For ASY-VRNet, separately encoded RGB and thermal features formed the visual input to its original visual-radar fusion, and projected returns were rasterized in the camera plane. CAFuser received RGB, thermal, and rasterized radar through modality-specific adapters while retaining its RGB-derived condition token. RADCI-RCINet retained its RGBT-radar concatenation and attention front end, followed by a class-agnostic mask decoder~\cite{guan2024asyvrnet,broedermann2025cafuser,radci}. We trained the adapted mask outputs with the same binary moving-instance targets and applied a common online Hungarian linker based on mask overlap and appearance similarity to methods without native identities. All compared methods used the same inputs and annotations.

\paragraph{Implementation details}
Training was performed in two stages. The detector was trained for 60 epochs with AdamW~\cite{loshchilov2019adamw}, an empty text prompt, and randomized static negatives with per-instance probability $0.5$ and box scale $1.5$. The tracker was trained for 1,000 updates on two RTX A6000 GPUs using bfloat16 and 32-frame clips split into four-frame graphs, with at most four identities per clip. The learning rates were $10^{-6}$ for the tracker and $2\!\times\!10^{-5}$ for the radar adapter, and radar dropout was $0.15$. The detector and radar encoder remained frozen during tracker training.


\paragraph{Metrics}
For segmentation, the IoU in Table~\ref{tab:main_results} is the macro-average of per-frame foreground overlap after merging all instances.  F1$_{50}$ evaluates instance recovery via Hungarian matching with an IoU$\geq 0.5$ correctness threshold.
For tracking, MOTA penalizes missed targets, false positives, and identity switches, whereas MOTP measures the localization quality of successfully matched predictions~\cite{bernardin2008clear}. HOTA balances detection accuracy with association accuracy~\cite{luiten2021hota}, while IDF1 is the identity-level F1 score and reflects how consistently detections retain the correct identity over time~\cite{ristani2016performance}. Higher is better for all reported metrics.
\subsection{Main Results}
Table~\ref{tab:main_results} reveals two gain regimes. Against the strongest competing score in each column, \method{} raises IoU from 0.6574 to 0.7027, an absolute gain of 0.0453, whereas F1$_{50}$ rises from 0.6243 to 0.8090, a gain of 0.1847. The larger F1$_{50}$ gain shows that the segmentation advantage is more pronounced for recovering matchable moving instances than for improving frame-averaged foreground overlap.

The contrast is larger for identity: MOTA, HOTA, and IDF1 improve by 0.2977, 0.1603, and 0.2857 over their respective strongest competing values. IoU is computed independently per frame, so preserving a track through an occlusion need not alter mask overlap on frames where the object is already visible. IDF1 instead measures correctly identified detections over a sequence, so fragmentation or reassignment can reduce the score across multiple frames. Radar association is designed to resolve ambiguous ownership, while Hold-Lost memory protects the last reliable state during radar-supported visual degradation. The fact that both components improve IDF1 in Table~\ref{tab:tracker_ablation} is consistent with this continuity-based explanation for the larger identity gain. Consistently, \method{} obtains 0.7559 MOTP, below ASY-VRNet at 0.7716, showing that the main benefit is temporal continuity rather than tighter localization of already matched masks. The multimodal baselines also do not uniformly outperform the RGB methods, showing that additional sensor inputs alone do not guarantee a gain on this benchmark.

\begin{figure*}[t]
    \centering
    \includegraphics[width=1.0\textwidth]{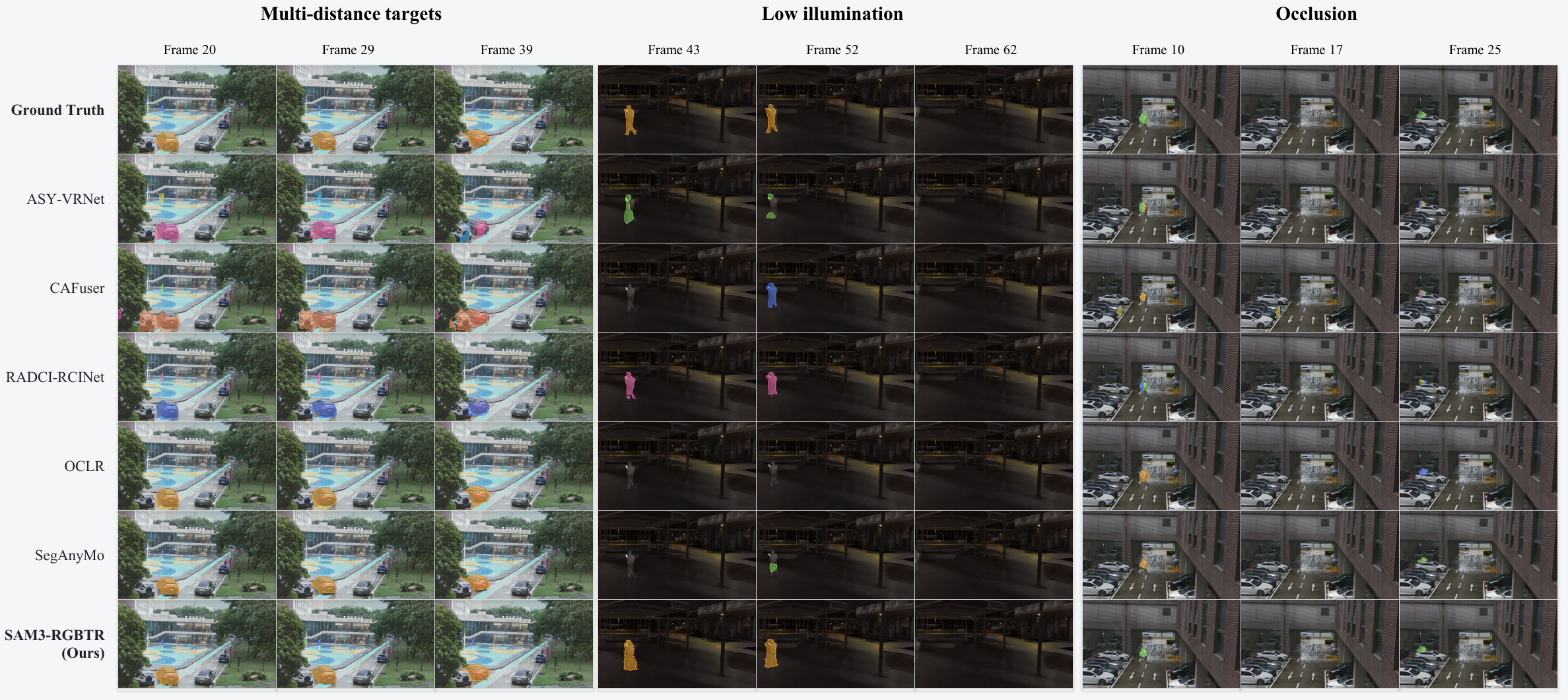}
    \caption{Qualitative comparison under multi-distance targets, low illumination, and occlusion. Frames progress from left to right within each condition; rows compare the ground truth, five baselines, and \method{}.}
    \label{fig:qualitative}
\end{figure*}

\subsection{Ablation Studies}
\paragraph{Detector components}

Table~\ref{tab:detector_ablation} reports the detector ablations. Adding motion supervision to radar attributes raises Pixel IoU from 0.4677 to 0.5850, Pixel Dice from 0.6373 to 0.7382, and F1$_{50}$ from 0.5599 to 0.7927. The largest absolute change is in F1$_{50}$, showing that return-level foreground classification improves instance recovery under the matching criterion, while the simultaneous pixel-metric gains show better foreground estimation. The full detector achieves the best overall values of 0.7540, 0.8598, and 0.8389.

\begin{table}[t]
    \centering
    \caption{Detector configurations and results. All rows include RGB; T, R, and M denote thermal fusion, radar attributes, and motion supervision. P-IoU and P-Dice are dataset-level pixel metrics.}
    \label{tab:detector_ablation}
    \setlength{\tabcolsep}{2pt}
    \begin{tabular}{@{}lcccccc@{}}
        \toprule
        Variant & T & R & M & P-IoU & P-Dice & F1$_{50}$ \\
        \midrule
        RGB only & - & - & - & 0.4064 & 0.5779 & 0.6517 \\
        Thermal fusion & $\checkmark$ & - & - & 0.4128 & 0.5844 & 0.7581 \\
        Radar attributes & - & $\checkmark$ & - & 0.4677 & 0.6373 & 0.5599 \\
        Motion supervision & - & $\checkmark$ & $\checkmark$ & 0.5850 & 0.7382 & 0.7927 \\
        \midrule
        Full detector & $\checkmark$ & $\checkmark$ & $\checkmark$ & 0.7540 & 0.8598 & 0.8389 \\
        \bottomrule
    \end{tabular}
\end{table}


\paragraph{Tracker components}
Table~\ref{tab:tracker_ablation} reports complementary gains from the two tracker mechanisms. Relative to the visual tracker, radar association raises MOTA/HOTA/IDF1 by 0.1291/0.0530/0.0903, while Hold-Lost memory raises them by 0.1190/0.0444/0.0755. Radar association is the stronger individual component, consistent with competitive ownership resolving which track should receive each return. Hold-Lost memory still raises IDF1 from 0.6233 to 0.6988, consistent with its role in protecting the stored identity state when visual evidence temporarily degrades. Combining the mechanisms gives the largest gains, 0.2229/0.1053/0.1380, and the best value for every reported metric. This result is consistent with their distinct roles: radar association selects the recipient track, whereas Hold-Lost memory decides whether an uncertain visual update may overwrite its state.


\begin{table}[t]
    \centering
    \caption{Tracker ablation. Radar association and Hold-Lost memory are each added to the visual tracker.}
    \label{tab:tracker_ablation}
    \begin{tabular}{lccc}
        \toprule
        Variant & MOTA & HOTA & IDF1 \\
        \midrule
        SAM~3 visual tracker  & 0.4007 & 0.4965 & 0.6233 \\
        SAM~3 + radar association  & 0.5298 & 0.5495 & 0.7136 \\
        SAM~3 + Hold-Lost memory  & 0.5197 & 0.5409 & 0.6988 \\
        Full tracker &0.6236 &0.6018 &0.7613 \\
        \bottomrule
    \end{tabular}
\end{table}

\subsection{Qualitative Results}

Figure~\ref{fig:qualitative} complements the aggregate metrics with three surveillance conditions. In the multi-distance sequence, \method{} follows the annotated movers with compact masks while several baselines produce fragmented or extraneous regions. Under low illumination, it recovers a coherent silhouette in Frames~43 and~52 and correctly produces no mask in Frame~62. In the occlusion sequence, it recovers the same track identity after the obstruction in Frames~10 and~25, whereas several competing rows miss the target or assign a different identity after it reappears. These examples mirror the two quantitative gain regimes: instance recovery improves F1$_{50}$, while continuity through degraded visual evidence improves IDF1 without requiring a comparable increase in framewise IoU.



%% file: sections/06_conclusion.tex
\section{Conclusion}
\label{sec:conclusion}
We introduced \benchmark{}, an RGBT-radar benchmark designed for motion segmentation and tracking under various surveillance conditions, together with \method{}, a framework that uses radar measurements to strengthen object discovery and temporal association. In the detector, a return's projected location identifies the image region it can support, while its range, radial velocity, and SNR provide physical motion and reliability cues beyond local RGBT appearance. Motion supervision then teaches the detector to distinguish returns associated with actual movers from multipath reflections and static clutter, allowing these radar attributes to contribute to more reliable instance predictions. During tracking, radar association connects each measurement to an individual trajectory, while Hold-Lost memory preserves the last reliable visual state when radar confirms that a visually degraded target remains present. The experiments consistently show that these designs improve instance recovery and identity continuity. These findings indicate that radar is most effective when its sparse measurements are explicitly grounded, assigned to individual tracks, and coupled to controlled visual memory. 
